\documentclass[letterpaper]{article} 
\usepackage{aaai24}  
\usepackage{times}  
\usepackage{helvet}  
\usepackage{subfig}
\usepackage{courier}  
\usepackage[hyphens]{url}  
\usepackage{graphicx} 
\usepackage{natbib}  
\usepackage{caption}
\usepackage{algorithm}
\usepackage{algorithmic}
\usepackage{multirow}

\usepackage{newfloat}
\usepackage{listings}
\DeclareCaptionStyle{ruled}{labelfont=normalfont,labelsep=colon,strut=off} 
\floatstyle{ruled}
\newfloat{listing}{tb}{lst}{}
\floatname{listing}{Listing}
\title{COPD-FlowNet: Elevating Non-invasive COPD Diagnosis with CFD Simulations (Student Abstract)} 
\author{
    Aryan Tyagi\textsuperscript{\rm *},
    Aryaman Rao\textsuperscript{\rm *},
    Shubhanshu Rao\textsuperscript{\rm },
    Raj Kumar Singh\textsuperscript{\rm }
}
\affiliations{
    Delhi Technological University, New Delhi, India\\

    \url{tyagiaryan82@gmail.com}

}

\usepackage{bibentry}

\begin{document}

\maketitle

\begin{abstract}
Chronic Obstructive Pulmonary Disorder (COPD) is a prevalent respiratory disease that significantly impacts the quality of life of affected individuals. This paper presents COPD-FlowNet, a novel deep-learning framework that leverages a custom Generative Adversarial Network (GAN) to generate synthetic Computational Fluid Dynamics (CFD) velocity flow field images specific to the trachea of COPD patients. These synthetic images serve as a valuable resource for data augmentation and model training. Additionally, COPD-FlowNet incorporates a custom Convolutional Neural Network (CNN) architecture to predict the location of the obstruction site. 
\end{abstract}

\section{Introduction}
Chronic Obstructive Pulmonary Disorder (COPD) is the third leading cause of death worldwide, causing over 3 million deaths in 2019 alone \cite{copd}. COPD is caused by smoking and exposure to air pollution that can hinder lung growth. Patients with COPD suffer from symptoms that include breathing difficulties and fatigue due to increased stiffness and blockage of the airways. Before starting treatment, it is imperative to accurately detect the location of the obstruction sites to devise a better treatment plan for the patient. Unfortunately, conventional diagnostic methods like bronchoscopy often prove impractical due to their invasiveness and high costs. Computational Fluid Dynamics (CFD) has emerged as a promising tool for airflow and blood flow simulations in patients \citep{ML1,ML2}, largely successful due to its non-invasive nature and cost-effectiveness. \citet{dataset} proposed a novel methodology for detecting the location of obstruction sites in COPD patients using CFD and Convolutional Neural Networks (CNN). However, it is not feasible to generate thousands of images using CFD simulations since they require a significant amount of computational resources. Moreover, datasets often suffer from imbalance, reflecting the prevalence of specific cases in the population. Thus, our study centers its efforts on augmenting and balancing the dataset created by \citet{dataset}. 

\def\thefootnote{*}\footnotetext{These authors contributed equally to this work.}\def\thefootnote{\arabic{footnote}}

\section{Methodology}
\subsection{Data Augmentation}
In the present study, a novel Generative Adversarial Network (COPD-FlowNet) is introduced to generate realistic velocity flow field images at a resolution of 128x128 pixels. The generator in this network employs a series of CNN layers integrated with BatchNorm layers and leaky-ReLU activation functions to achieve the desired image dimensions. Meanwhile, the discriminator incorporates traditional Conv2D layers, along with fully connected layers, including Dropout and Dense layers, featuring several innovative adaptations. 

\subsection{Classification}
A custom CNN network for classifying the location of obstruction sites in the patient's lungs is also introduced. The proposed classifier takes an input image of 128x128 pixels as the input layer. It is followed by a sequence of convolution blocks, with the corresponding depth continuously increasing from 8 filters to 128 filters. With each convolution block,  the kernel size is decreased, which allows the network to capture finer details in the data. The padding is set to "valid," which ensures that the size of the output feature maps is reduced. The ReLU activation function introduces non-linearity, while MaxPooling2D layers and batch normalization enhance feature extraction and model convergence. The Batch normalization is inserted to mitigate the "covariate shift" problem, thereby enhancing convergence during training. 

\begin{figure*}[h]
    \centering
    \includegraphics[scale=0.25]{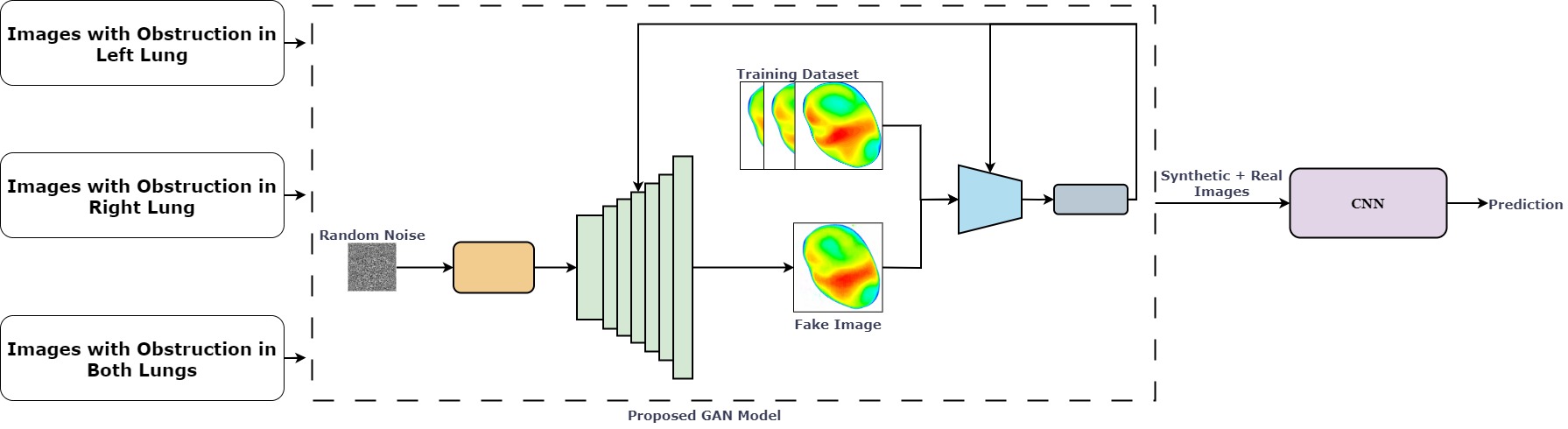}
    \caption{Overall Architecture of COPD-FlowNet}
    \label{flowchart}
\end{figure*}

\begin{table}[h!]
    \centering
    \begin{tabular}{c||c|c|c|c}
         \hline\hline
         Dataset Distribution & Left & Right & Both & Total \\
         \hline
         Original Dataset & 45 & 113 & 137 & 295 \\
         After Augmentation & 400 & 400 & 400 & 1200 \\ 
         \hline\hline
    \end{tabular}
    \caption{Distribution of dataset}
    \label{distribution}
\end{table}

\section{Experimental Setup}
The dataset used consisted of 295 images of velocity flow fields at the trachea of patients with COPD. These images were obtained by CFD simulations of airflow in patient-specific lung geometries. The dataset is divided into three classes named left, right, and both, depending on the location of the obstruction in the lung. The exact distribution of the dataset before and after augmentation is given in Table \ref{distribution}. 

\section{Experiment Analysis and Results}
\begin{figure}[]
  \centering
  \subfloat[]{\includegraphics[width=0.22\textwidth]{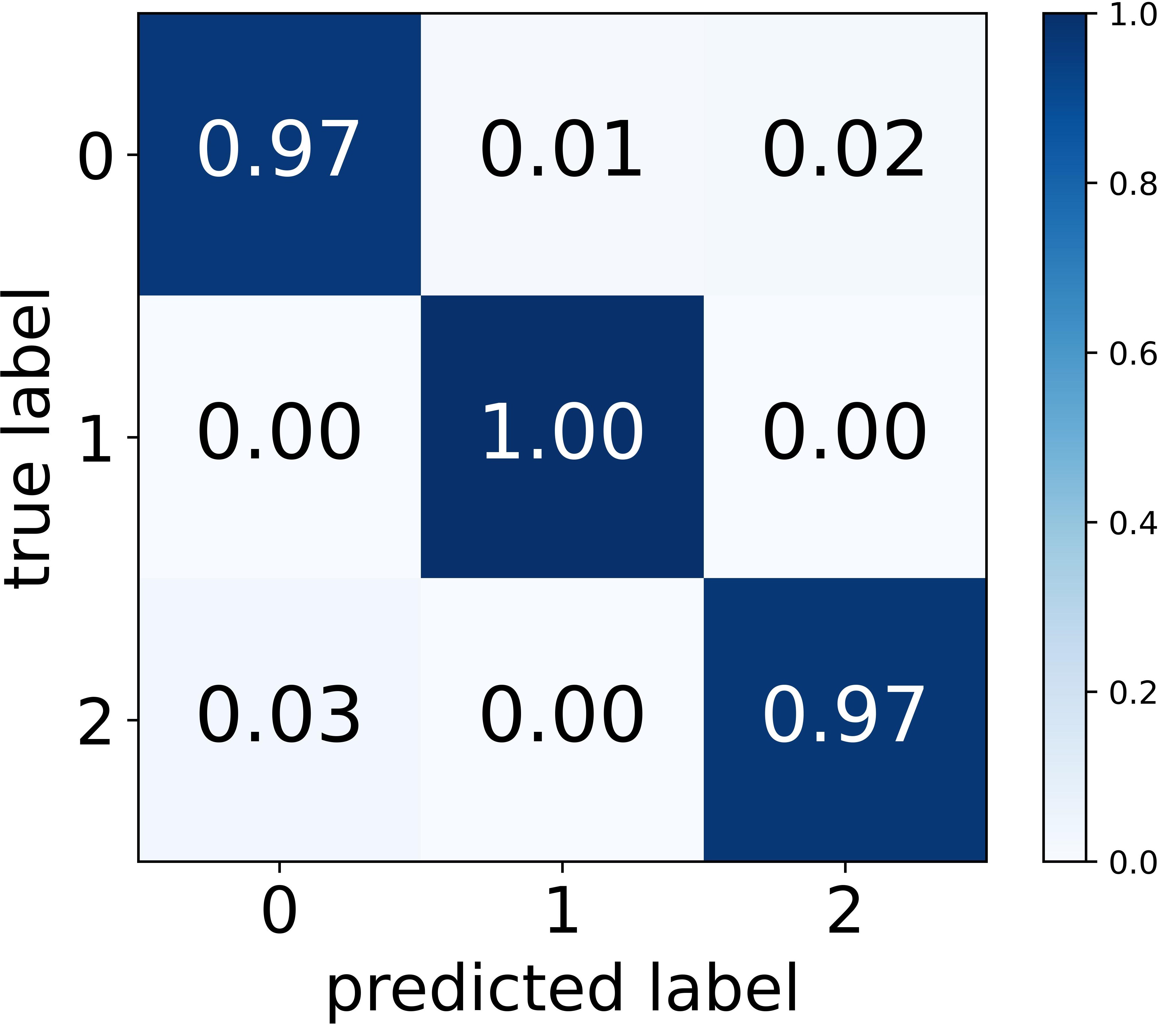}\label{fig:f1}}
  \hfill
  \subfloat[]{\includegraphics[width=0.25\textwidth]{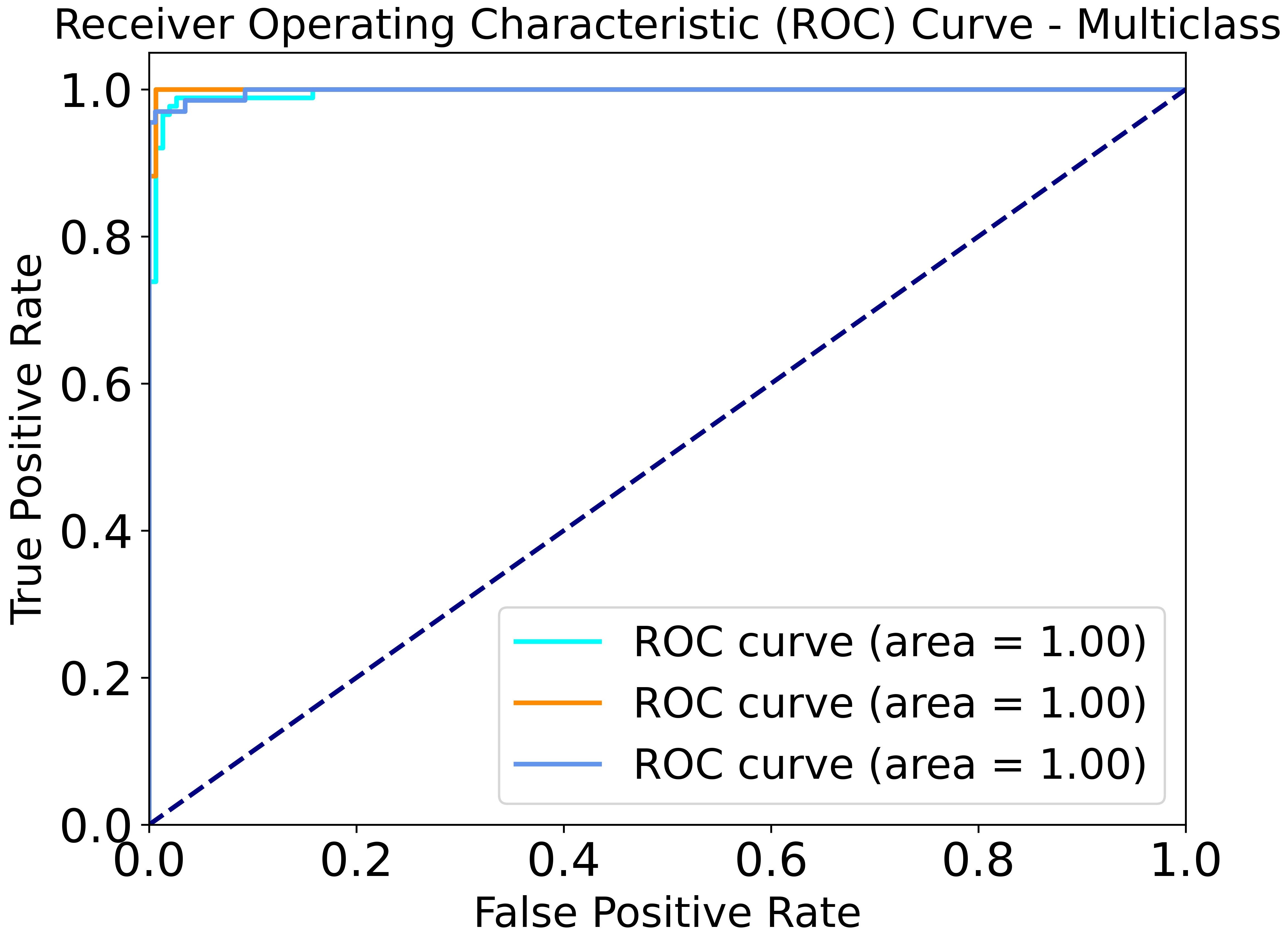}\label{fig:f2}}
  \caption{(a) Normalized confusion matrix, (b) Receiver Operating Characteristic (ROC) curve}
  \label{resultsFig}
\end{figure}

Our classification model outperforms well-established benchmark algorithms such as ResNet-50 and EfficientNet.  Notably, before augmentation, our CNN model achieved an accuracy of 73.94\%. However, after implementing augmentation techniques, this accuracy increased to 96.35\%. Consequently, our approach surpassed the performance of state-of-the-art (SOTA) algorithms in the conventional performance metrics such as Accuracy and F1-score, as illustrated in Table \ref{table:1}. The normalized confusion matrix and ROC curve displayed in Figure \ref{resultsFig}, depict the predicted test sample distribution towards each class and the classification model's performance at different classification thresholds respectively.

\begin{table}[h!]
\centering 
\scalebox{0.83}{
\begin{tabular}{ c || c| c| c| c| c} 
 \hline\hline  
 Method & Accuracy & Class & Precision & Recall & F1 Score\\ 
 \hline 
 \multirow{3}{*}{Proposed} & \multirow{3}{*}{\textbf{96.35\%}} & Left & 0.98 & 0.92 & 0.95  \\ 
 & & Right & 0.97 & 1.00 & 0.98  \\ 
 & & Both & 0.94 & 0.97 & 0.96  \\ 
 \hline
 \multirow{3}{*}{ResNet-50} & \multirow{3}{*}{96\%} & Left & 0.94 & 0.94 & 0.94  \\ 
 & & Right & 0.96 & 0.98 & 0.97  \\ 
 & & Both & 0.98 & 0.97 & 0.97  \\ [0.5ex]
 \hline
 \multirow{3}{*}{EfficientNet} & \multirow{3}{*}{95.00\%} & Left & 0.94 & 0.92 & 0.93  \\ 
 & & Right & 0.96 & 0.96 & 0.96  \\ 
 & & Both & 0.94 & 0.97 & 0.95  \\ [0.5ex] 
 \hline\hline
\end{tabular}
}

\caption{Experimental Analysis}
\label{table:1}
\end{table}


\section{Conclusion}
Our proposed approach represents an integration of deep learning and CFD. It demonstrates the potential use of CFD for medical diagnosis and its enhancement using generative networks. Further, the proposed CNN outperforms well-established models such as ResNet-50 and EfficientNet. Future works will include generating datasets that can detect more specific locations of the obstruction sites since currently the model only predicts whether the obstruction is in the left lung, right lung, or both lungs.

\bibliography{aaai24}

\end{document}